\documentclass{article}
\usepackage{ijcai25}

\usepackage{times}
\usepackage{soul}
\usepackage{url}
\usepackage[hidelinks]{hyperref}
\usepackage[utf8]{inputenc}
\usepackage[small]{caption}
\usepackage{graphicx}
\usepackage{amsmath}
\usepackage{amsthm}
\usepackage{booktabs}
\usepackage{algorithm}
\usepackage{algorithmic}
\usepackage[switch]{lineno}
\usepackage{amsmath}
\usepackage{amssymb}
\usepackage{multirow}
\usepackage{xcolor}  
\usepackage{colortbl} 
\usepackage{graphicx} 

\title{MSCI: Addressing CLIP's Inherent Limitations for Compositional Zero-Shot Learning}

\author{
Yue Wang$^{1,\dag}$
\and
Shuai Xu$^{1,2,\dag,*}$\and
Xuelin Zhu$^{3}$\And
Yicong Li$^1$\\
\affiliations
$^1$Nanjing University of Aeronautics and Astronautics, Nanjing, China\\
$^2$Key Laboratory of Social Computing and Cognitive Intelligence (Dalian University of Technology), Ministry of Education, China\\
$^3$The Hong Kong Polytechnic University, Hong Kong, China\\
\emails
\{wangyue11, xushuai7\}@nuaa.edu.cn,
zhuxuelin23@gmail.com,
liyicong@nuaa.edu.cn
}

\begin{document}

\maketitle
\begingroup
\renewcommand\thefootnote{}\footnote{
\hspace{-1em}$^\dag$ Equal contribution.
$^*$ Corresponding author.
}
\addtocounter{footnote}{-1}
\endgroup

\begin{abstract}
Compositional Zero-Shot Learning (CZSL) aims to recognize unseen state-object combinations by leveraging known combinations. Existing studies basically rely on the cross-modal alignment capabilities of CLIP but tend to overlook its limitations in capturing fine-grained local features, which arise from its architectural and training paradigm. To address this issue, we propose a \textbf{M}ulti-\textbf{S}tage \textbf{C}ross-modal \textbf{I}nteraction (MSCI) model that effectively explores and utilizes intermediate-layer information from CLIP's visual encoder. Specifically, we design two self-adaptive aggregators to extract local information from low-level visual features and integrate global information from high-level visual features, respectively. These key information are progressively incorporated into textual representations through a stage-by-stage interaction mechanism, significantly enhancing the model’s perception capability for fine-grained local visual information. Additionally, MSCI dynamically adjusts the attention weights between global and local visual information based on different combinations, as well as different elements within the same combination, allowing it to flexibly adapt to diverse scenarios. Experiments on three widely used datasets fully validate the effectiveness and superiority of the proposed model. Data and code are available at \textcolor{blue}{\url{https://github.com/ltpwy/MSCI}}.
\end{abstract}

\section{Introduction}

Compositional Zero-Shot Learning (CZSL) \cite{r32} aims to strategically disassemble and recompose visual representations of seen combinations (composed of a state and an object, such as ``tall building" or ``green tree") to construct representations of new composite classes (e.g., ``tall tree"), thereby enabling precise recognition of them.

\begin{figure}[htbp] 
    \centering 
    \includegraphics[width=0.39\textwidth]{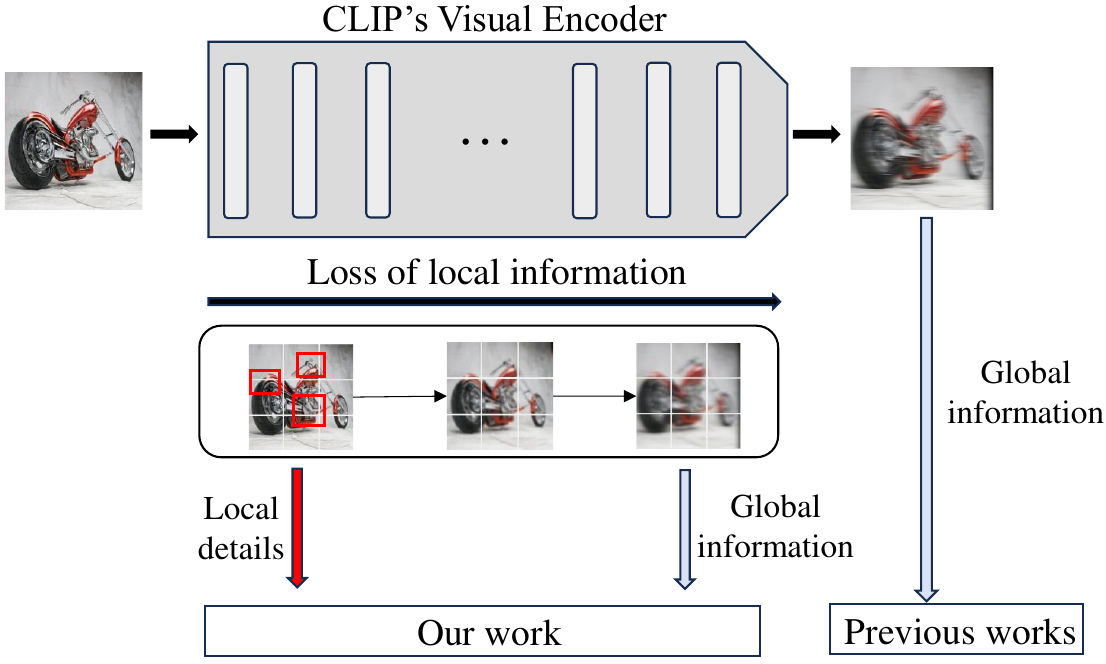} 
    \caption{Core idea of our work. By effectively leveraging the rich local details in the lower-level features of the visual encoder, CLIP's ability to capture fine-grained local information can be enhanced.} 
    \label{fig:1} 
\end{figure}

In the early research of CZSL, there is a greater focus on how to effectively integrate and leverage existing visual information to recognize unseen categories. Some methods treat state-object pairs as a single entity, directly learning their compatibility feature representations with images \cite{r1,r2}. Furthermore, researches \cite{r3,r4} have attempted to explicitly separate attributes and objects through spatial embedding techniques to optimize their combination process. However, due to the absence of a unified feature space and effective attribute-object decoupling modeling, these methods struggle with cross-modal alignment, significantly limiting the model's performance.

The invention of CLIP \cite{r5} effectively addresses the challenge of cross-modal alignment. Using large-scale pretraining data and a contrastive learning strategy, CLIP demonstrates strong cross-modal alignment capabilities, which has led to the emergence of numerous methods that apply CLIP to downstream CZSL tasks. Zhou et al. \cite{r6} are the first to combine CLIP with prompt engineering, proposing a single-path joint training paradigm in which the text embeddings of state-object pairs generated by CLIP are used as parameters and updated during backpropagation. Nayak et al. \cite{r7} further refine this approach by introducing adjustable vocabulary tokens to represent primitive concepts in a compositional manner. Huang et al. \cite{r8} propose an innovative multi-path paradigm, training decouplers to disentangle visual features and interact the disentangled features separately with corresponding prompt embeddings. Jing et al. \cite{r9} strengthen disentangled representations of states and objects by exploring internal connections between the same combination of objects and the same combination of states.

These methods fully leverage the powerful cross-modal alignment capabilities of CLIP, achieving remarkable results. However, they generally overlook the inherent limitations of CLIP itself. From a local perspective, the CLIP image encoder, based on a transformer architecture, compresses the entire image into a fixed global feature vector. To improve computational efficiency and training speed, the model tends to focus on global visual information, while being less sensitive to fine-grained local details. From a global perspective, CLIP’s contrastive learning objective aims to maximize the similarity between global features of matched image-text pairs while minimizing the similarity of mismatched pairs. This global optimization strategy prioritizes capturing the overall alignment between images and texts in the semantic space, rather than aligning fine-grained local features. Consequently, for tasks requiring precise differentiation of local features, such as fine-grained modeling of state-object combinations in CZSL, CLIP's performance is often limited.

Therefore, to address the aforementioned issues, this paper proposes MSCI, a \textbf{M}ulti-\textbf{S}tage \textbf{C}ross-modal \textbf{I}nteraction model for compositional zero-shot image classification. The model fully leverages CLIP's strengths in cross-modal alignment while compensating for its shortcomings in handling fine-grained local features, as illustrated in Figure \ref{fig:1}. Unlike previous CZSL models that rely solely on the features of the output layer, MSCI employs two trainable feature aggregators to extract local visual information and global visual information from low- and high-level visual features, respectively. By interacting with textual embeddings in a stage-by-stage manner, MSCI not only integrates global visual information into the text features, but also captures valuable local details that are often overlooked, thereby significantly enhancing the model's accuracy and generalization capability in recognizing unseen combinations.

Furthermore, to enable the model to dynamically adjust its focus on local and global visual information based on different combinations, as well as the different elements within the same combination (i.e., states and objects), we propose a fusion module to regulate the relative influence of local and global visual features on the final text embeddings. This mechanism greatly enhances the model's ability to handle complex tasks and improves its adaptability to a wide range of scenarios.

The contributions of this paper are summarized as follows: 
\begin{itemize}

 \item We are the first to emphasize CLIP's inherent limitations in local feature perception for CZSL tasks due to its architecture and training paradigm, and propose addressing this issue by effectively utilizing the intermediate layer information of its visual encoder.
 \item We propose the MSCI model for compositional zero-shot learning. Through stage-wise feature fusion and interaction, we progressively enhance the relationships between text embeddings, local visual information, and global visual information, ensuring their collaborative interaction in cross-modal tasks.

 \item We validate the effectiveness of the proposed model through experiments and the results show that the model achieves state-of-the-art performance on the majority of key metrics across three widely used datasets under both open-world and closed-world settings.

\end{itemize}

\section{Related Work}

\subsection{Compositional Zero-Shot Learning}

CZSL is a specialized form of zero-shot learning that does not rely on any auxiliary information. Its core objective is to achieve generalization from known combinations to unseen combinations by decoupling and recombining visual features. Current CZSL models can be broadly categorized into two types: CLIP-based models and non-CLIP-based models.

In CLIP-based CZSL models, Zhou et al. \cite{r6} were the first to propose combining prompt engineering with pre-trained vision language models (VLM) to address efficiency issues when designing prompts for downstream tasks. By learning adjustable contextual word vectors, they achieved automated generation of prompt sentences, effectively mitigating the dependency on task-specific prompt designs. To address the limitations of VLMs in downstream CZSL tasks, Nayak et al. \cite{r7} treated attribute and object tokens defining categories as learnable parameters, optimizing them through multiple combinations of prompts. Xu et al. \cite{r10} further modeled the compositional relationships between objects and attributes as graph structures, treating attributes and object labels as graph nodes, and utilized Graph Neural Networks (GNNs) \cite{r11,r34} to update and optimize soft prompt representations.

In particular, Huang et al. \cite{r8} extended the single-path paradigm to a multi-path framework, establishing separate recognition branches for states, objects, and their combinations. They also introduced a cross-modal alignment module to better align prompt representations with current visual content. However, this approach focuses only on the interaction between final-layer visual features and text, resulting in the loss of significant local information during the forward pass of the visual encoder. Building on the multi-path paradigm, Jing et al. \cite{r9} further enhanced the decoupling of visual features by constructing a database of related samples.

Although these methods have made significant progress in adapting CLIP's cross-modal alignment capabilities to CZSL tasks, they often overlook the inherent limitations of CLIP in its architecture and contrastive learning training paradigm, specifically its weaker sensitivity to fine-grained local features. In contrast, the proposed MSCI model directly addresses this critical limitation, providing a more robust and higher-performing solution for CZSL tasks.

\begin{figure*}[htbp] 
    \centering 
    \includegraphics[width=0.92\textwidth]{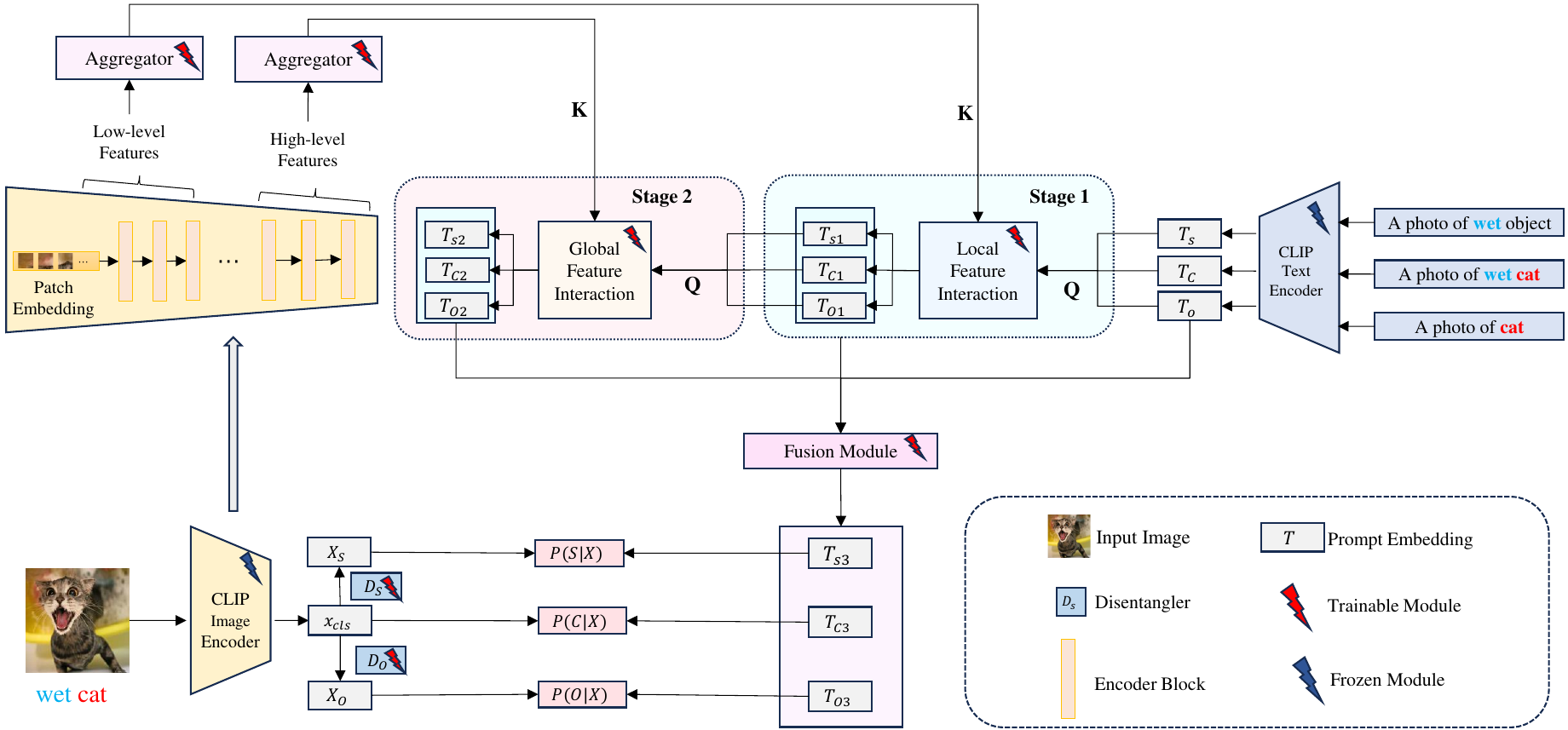} 
    \caption{The general framework of the MSCI model. We design two self-adaptive aggregators to extract local and global visual information from low-level and high-level visual features, respectively, and integrate this information into the prompt embeddings through stage-wise cross-modal interactions. Additionally, a fusion module is introduced to regulate the influence of local and global information on the generation of the final prompt embeddings.} 
    \label{fig:2} 
\end{figure*}

\subsection{Multi-layer Feature Aggregation}
In recent years, the exploration of intermediate-layer information in Transformers for downstream tasks has garnered significant attention in the field of computer vision. By leveraging the multi-level and multi-scale feature information contained in intermediate layers, this approach effectively addresses the limitations of traditional deep learning models that rely solely on high-level features. For example, Tang et al. \cite{r12} utilized the boundary characteristics of low-level features and semantic information of high-level features, applying them to medical image segmentation tasks. Similarly, Liu et al. \cite{r13} enhanced the capture of multi-scale local details and structural relationships by jointly training multi-layer feature learning and encoding modules with Transformers, achieving outstanding performance in malicious webpage detection. Furthermore, this idea has been extended to other domains \cite{r37,r35}, such as cross-modal retrieval \cite{r14,r38} and visual localization \cite{r15,r36}, demonstrating its broad applicability.

\section{Methodology}
This section begins by providing a formal definition of the CZSL task, which serves as the foundation for analyzing the inherent limitations of CLIP when tackling downstream CZSL tasks. Building on this analysis, we present our proposed model in detail. The core of the model lies in aggregating multi-layer information from CLIP's visual encoder and conducting stage-wise cross-modal interactions with textual embeddings. This design enables the model to precisely integrate global visual information with local visual features, facilitating adaptive adjustment of prompt representations. By doing so, our model effectively addresses CLIP's limitations in perceiving fine-grained local features. The general framework of the proposed model is illustrated in Figure \ref{fig:2}.

\subsection{Preliminaries}
\subsubsection{Problem Formulation}
Given a state set \( S = \{s_0, s_1, \dots, s_n\} \) and an object set \( O = \{o_0, o_1, \dots, o_m\} \), a label space \( C \) can be constructed via the Cartesian product, denoted as \( C = S \times O \). From \( C \), we extract two non-intersecting subsets: the seen class set \( C_s \) and the unseen class set \( C_u \), satisfying \( C_s \cup C_u \subseteq C \) and \( C_s \cap C_u = \emptyset \). During the training phase, the task of CZSL is to learn a discriminative mapping \( P: X \to C_s \) from the input image space \( X \) to \( C_s \). In the testing phase, given an image \( I \), the task is to predict a class label \( c = (s, o) \) from the test class set \( C_{test} \) using the learned discriminative mapping \( P \).

Depending on the search space, the CZSL tasks are configured in two settings: In the closed-world setting, only the predefined combination space is considered, i.e., \( C_{test} = C_s \cup C_u \); In the more challenging open-world setting, the search space includes all possible pairs of state objects, i.e., \( C_{test} = C \).

\subsubsection{Limitations of CLIP}
The limitations of CLIP in local feature perception can be primarily attributed to two factors: the design of its visual encoder architecture and its contrastive learning-based training paradigm. CLIP's visual encoder is based on the Transformer architecture, which excels in modeling long-range feature dependencies through its global attention mechanism at the expense of sacrificing local details. This limitation is particularly evident in its sub-par performance when capturing low-level features such as edges and textures. Furthermore, the CLIP training objective aims to maximize global semantic alignment between images and texts through contrastive learning, causing the model to prioritize capturing general semantic information while neglecting finer local details. Additionally, the contrastive learning paradigm requires the model to rapidly distinguish prominent features across images, further diminishing its sensitivity to fine-grained local features.

\subsubsection{Feature Encoding}
We use the CLIP image encoder as the visual backbone, which is based on the ViT-L/14 architecture. For an input image \(I\) from the image set \(X\), we extract the \([CLS]\) token \(\boldsymbol{I}_{cls}\) from the output layer as its embedding representation. Building on this, we follow the three-path paradigm from previous work, where the image embedding \(\boldsymbol{I}_{cls}\) serves as the input to three independent multi-layer perceptrons (MLPs) \cite{r30} to generate visual representations of combinations, states and objects, which are denoted as \(\boldsymbol{V}_{com}\),\(\boldsymbol{V}_{state}\), \(\boldsymbol{V}_{obj}\), respectively. At the text level, we design soft prompt templates in the following forms: “a photo of \([state]\) \([object]\)”, “a photo of \([state]\) object” and “a photo of \([object]\)”, which are used to construct prompts for all candidate combinations, states, and objects, respectively. These prompts are then fed into the CLIP text encoder to generate prompt embeddings \(\boldsymbol{t}_{com}\), \(\boldsymbol{t}_{state}\), and \(\boldsymbol{t}_{obj}\). Their dimensions are \([N_{com}, d]\), \([N_{state}, d]\), and \([N_{obj}, d]\), where \(N_{com}\), \(N_{state}\), and \(N_{obj}\) denote the numbers of all candidate combinations, states and objects, respectively, and \(d\) represents the embedding dimension. We treat the embeddings of \([state]\) and \([object]\) as trainable parameters for fine-tuning.

\subsection{Aggregation of Multi-layer Information}
\begin{figure}[t] 
    \centering 
    \includegraphics[width=0.45\textwidth]{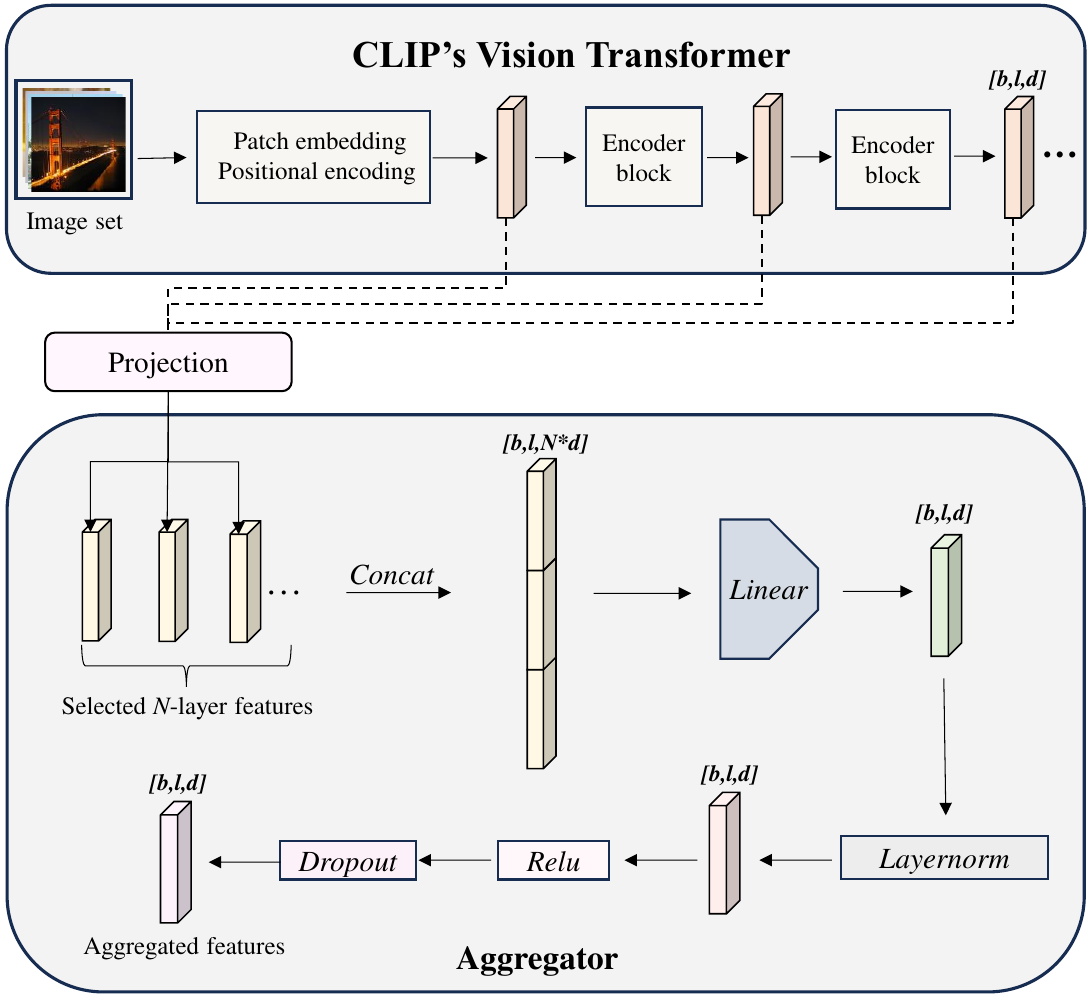} 
    \caption{Illustration of the low-level feature aggregator.} 
    \label{fig:3} 
\end{figure}

In the ViT architecture employed by the CLIP visual encoder, features at different levels exhibit unique information characteristics: the lower layers contain rich local detail information of the image, while the higher layers tend to integrate global structural features. To effectively utilize the information between layers, we design a self-adaptive feature aggregation module, as shown in Figure \ref{fig:3}. 

Suppose that the visual feature of the \(i\)-th layer are denoted as \(\boldsymbol{F}_i\), with dimensions \([b, l, d]\), where \(b\) is the number of images in the set of images \(X\), \(l\) represents the number of patches generated after convolution (including the token \([CLS]\)). We extract the features from the first \(N\) layers and the last \(M\) layers of the CLIP visual encoder, which are then concatenated separately along the feature dimension to form a richer feature representation. The concatenated features can be expressed as:
\begin{equation}
\boldsymbol{F}_{first\_n} = \text{Concat}(\boldsymbol{F}_1, \boldsymbol{F}_2, \dots, \boldsymbol{F}_N)
\end{equation}

\begin{equation}
\boldsymbol{F}_{last\_m} = \text{Concat}(\boldsymbol{F}_{S-M+1}, \boldsymbol{F}_{S-M+2}, \dots, \boldsymbol{F}_S)
\end{equation}

\noindent where \( S \) is the total number of encoder blocks in the CLIP visual encoder, \(\boldsymbol{F}_{first\_n} \) and \(\boldsymbol{F}_{last\_m} \) represent the concatenation of the first \(N\) and last \(M\) layers' features, with dimensions \([b, l, N \times d]\) and \([b, l, M \times d]\), respectively. The concatenated features are first passed through a linear transformation, mapping them from the concatenated dimension (\(N(M) \times d\)) to the target feature dimension \(d\), followed by layer normalization to ensure training stability. Then, ReLU activation is applied to introduce non-linearity, enhancing the model's ability to capture complex feature relationships. Finally, a Dropout layer is used to improve the model's generalization capability. The final fused low-level and high-level features, \( \boldsymbol{F}_{low} \) and \( \boldsymbol{F}_{high} \), can be expressed as:

\begin{equation}
\boldsymbol{F}_{low} = \mathcal{D}\left(\max\left(0, \frac{\boldsymbol{W} \boldsymbol{F}_{first\_n} + \boldsymbol{b} - \boldsymbol{\mu}}{\boldsymbol{\sigma}}\right), p \right)
\end{equation}

\begin{equation}
\boldsymbol{F}_{high} = \mathcal{D}\left(\max\left(0, \frac{\boldsymbol{W} \boldsymbol{F}_{last\_m} + \boldsymbol{b} - \boldsymbol{\mu}}{\boldsymbol{\sigma}}\right), p \right)
\end{equation}

\noindent where \( \boldsymbol{W} \in \mathbb {R}^{d \times (N(M) \times d)} \) is the weight matrix for the linear transformation, \( \boldsymbol{b} \in \mathbb{R}^{d} \) is the bias term, \( \boldsymbol{\mu} \) and \( \boldsymbol{\sigma} \) are the mean and standard deviation of the output features after the fully connected layer, and \( \mathcal{D}(\cdot, p) \) represents the Dropout operation with \( p \) as the dropout probability.

\subsection{Multi-stage Cross-modal Interaction}
The fused low-level features \( \boldsymbol{F}_{low} \) capture rich local visual details, while the fused high-level features \( \boldsymbol{F}_{high} \) integrate more abstract global visual information. They interact in a stage-wise manner with the prompt embedding \( \boldsymbol{t} \) from any branch, that is, \( \boldsymbol{t} \) can be any of \( \boldsymbol{t}_{com} \), \( \boldsymbol{t}_{state} \), or \( \boldsymbol{t}_{obj} \).

In the first stage, the prompt embedding \(\boldsymbol{t} \) interacts cross-modally with the fused low-level features \( \boldsymbol{F}_{low} \) to effectively integrate the rich local details contained in the low-level features into the prompt embedding. This interaction is achieved through a cross-attention layer combined with a residual connection, as represented by:
\begin{equation}
\begin{aligned}
\boldsymbol{t}' &= \text{CrossAttention}(\boldsymbol{t}, \boldsymbol{F}_{low}, \boldsymbol{F}_{low}) + \boldsymbol{t} \\
   &= \text{Softmax}\left( \frac{\boldsymbol{t} \boldsymbol{F}_{low}^T}{\sqrt{d}} \right) \boldsymbol{F}_{low} + \boldsymbol{t}
\end{aligned}
\end{equation}

\noindent where \(\boldsymbol{t'} \) denotes the updated prompt embedding after the cross-modal interaction and \(d\) is the dimension of attention.

Furthermore, we adopt the Feed-Forward Network (FFN) design proposed by Huang et al.\cite{r8}, which is implemented through an MLP. This network aims to optimize the feature representations after interaction and generates the output by combining the residual connection, as represented by:
\begin{equation}
\boldsymbol{t}_1 = \text{MLP}(\boldsymbol{t}') + \boldsymbol{t}'
\end{equation}

\noindent where \(\boldsymbol{t}_1\) denotes the updated prompt embedding after the FFN. After the first stage, prompt embedding integrates rich local visual information from low-level features. 

The second stage adopts an interaction pattern similar to that in the first stage, with the aim of further integrating the more abstract global visual information contained in the high-level features into the prompt embedding. We use the prompt embedding \(\boldsymbol{t}_1\) obtained from the first stage and the fused high-level visual features \(\boldsymbol{ F}_{high} \) as input. These are processed through a cross-attention layer and a feed-forward network, resulting in the further update of the prompt embedding. This process is represented as follows:
\begin{equation}
\begin{aligned}
\boldsymbol{t}'' &= \text{CrossAttention}(\boldsymbol{t}_1, \boldsymbol{F}_{high}, \boldsymbol{F}_{high}) + \boldsymbol{t}_1 \\
   &= \text{Softmax}\left( \frac{\boldsymbol{t} \boldsymbol{F}_{high}^T}{\sqrt{d}} \right) \boldsymbol{F}_{high} + \boldsymbol{t}_1
\end{aligned}
\end{equation}

\begin{equation}
\boldsymbol{t}_2 = \text{MLP}(\boldsymbol{t}'') + \boldsymbol{t}''
\end{equation}

Compared with \(\boldsymbol{t}_1 \), \(\boldsymbol{t}_2 \) further integrates the abstract global visual information contained in the high-level visual features. To dynamically assign attention weights to local and global visual information based on different combinations, as well as the different prompt branches of the same combination, we introduce two learnable parameters \(\lambda_1\) and \(\lambda_2\) to regulate the weights of \( \boldsymbol{t}_1 \) and \(\boldsymbol{t}_2 \) in the final prompt embedding. The final prompt embedding is represented as follows:

\begin{equation}
\boldsymbol{t}_{final} = \boldsymbol{t} + \lambda_1 \boldsymbol{t}_1 + \lambda_2 \boldsymbol{t}_2
\end{equation}

\subsection{Training and Inference}
We follow the standard training and inference process of the multi-path paradigm. Assume that the initial prompt embeddings \( \boldsymbol{t}_{com} \), \( \boldsymbol{t}_{state} \), and \( \boldsymbol{t}_{obj} \) are transformed into \( \boldsymbol{T}_{com} \), \( \boldsymbol{T}_{state} \), and \( \boldsymbol{T}_{obj} \) through multi-stage interactions. The probability of assigning the combination label \( \boldsymbol{c}(s, o) \), the state label \( \boldsymbol{s} \), and the object label \( \boldsymbol{o} \) to the image \( I \) can be expressed as:

\begin{equation}
p(c | I) = \frac{\exp(\boldsymbol{V}_{com} \cdot \boldsymbol{T}_{com}^c / \tau)}{\sum_{k=1}^{N_{com}} \exp(\boldsymbol{V}_{com} \cdot \boldsymbol{T}_{com}^k / \tau)},
\end{equation}

\begin{equation}
p(s | I) = \frac{\exp(\boldsymbol{V}_{state} \cdot \boldsymbol{T}_{state}^s / \tau)}{\sum_{k=1}^{N_{state}} \exp(\boldsymbol{V}_{state} \cdot \boldsymbol{T}_{state}^k / \tau)},
\end{equation}

\begin{equation}
p(o | I) = \frac{\exp(\boldsymbol{N}_{obj} \cdot \boldsymbol{T}_{obj}^o / \tau)}{\sum_{k=1}^{N_{obj}} \exp(\boldsymbol{T}_{obj} \cdot \boldsymbol{V}_{obj}^k / \tau)},
\end{equation}

\noindent 
where \( \tau \in \mathbb{R} \) represents the pre-trained temperature parameter, \( \boldsymbol{T}_{com}^c \), \( \boldsymbol{T}_{state}^s \), and \( \boldsymbol{T}_{obj}^o \) represent the prompt embeddings of the combination \( c \), the state \( s \), and the object \( o \), respectively. The probabilities predicted by each branch are compared with the one-hot encoded labels using cross-entropy to compute the loss. The total training loss is then obtained as a weighted sum of the losses from each branch, formulated as follows:

\begin{equation}
L_s = - \frac{1}{|X|} \sum_{x \in X} \log p(s | x)
\end{equation}

\begin{equation}
L_o = - \frac{1}{|X|} \sum_{x \in X} \log p(o | x)
\end{equation}

\begin{equation}
L_c = - \frac{1}{|X|} \sum_{x \in X} \log p(c | x)
\end{equation}

\begin{equation}
L_{\text{total}} = \alpha_s L_s + \alpha_o L_o + \alpha_c L_c
\end{equation}

In the inference phase, for an input image \( A \), suppose that \( C(s_i, o_j) \) is an arbitrary combination in the search space \( S \). The model predicts the most likely combination \(\hat{c}\) based on the following formula:

\begin{equation}
\hat{c} = \arg\hspace{-0.8em}\max_{\raisebox{-0.6ex}{\(\scriptstyle C(s_i, o_j) \in S\)}}\hspace{-0.8em} 
\beta \left( p(C(s_i, o_j) |A ) \right) + (1 - \beta) \cdot p(s_i | A) \cdot p(o_j | A)
\end{equation}

\noindent where \(\beta\) is a predefined parameter used to control the proportion of each branch's results during the inference process.

\section{Experiment}

\subsection{Experimental Setup}
\subsubsection{Datasets} 
We evaluate the performance of the proposed MSCI on three widely-used compositional zero-shot learning datasets: MIT-States \cite{r16}, UT-Zappos \cite{r17}, and C-GQA \cite{r2}. The MIT-States dataset contains 53,753 images involving 245 object categories and 115 state categories. The UT-Zappos dataset includes 50,025 images, covering 12 object categories and 16 state categories. C-GQA, constructed based on the GQA dataset \cite{r33}, comprises 870 object categories and 453 state categories. Consistent with previous research, we adopt the dataset partitioning method proposed by Purushwalkam et al. \cite{r1} , with specific details presented in Table \ref{tab:1}.

\begin{table}[ht]
\centering

\resizebox{\linewidth}{!}{%
\begin{tabular}{@{}lcccccccc@{}}
\toprule
\textbf{Dataset} &  \multicolumn{2}{c}{\textbf{Train}}  & \multicolumn{3}{c}{\textbf{Validation}} & \multicolumn{3}{c}{\textbf{Test}} \\
\cmidrule(lr){2-3} \cmidrule(lr){4-6} \cmidrule(lr){7-9}
 & $|Y_s|$ & $|X|$& $|Y_s|$ & $|Y_u|$ & $|X|$ & $|Y_s|$ & $|Y_u|$ & $|X|$ \\
\midrule
MIT-States & 1,262&30,338 & 300 & 300 & 10,420 & 400 & 400 & 12,995 \\
UT-Zappos  & 83&22,998 & 15  & 15  & 3,214  & 18  & 18  & 2,914  \\
C-GQA      & 5,592& 26,920 & 1,252 & 1,040 & 7,280  & 888 & 923 & 5,098 \\
\bottomrule
\end{tabular}%
}
\caption{Statistics of datasets}
\label{tab:1}
\end{table}

\subsubsection{Metrics}
We follow the standard evaluation protocols adopted in previous studies \cite{r7} to comprehensively evaluate the performance of the model in both close- and open-world settings. Specifically, the evaluation metrics include the best seen accuracy (S), the best unseen accuracy (U), the best harmonic mean (HM), and the area under the seen-unseen accuracy curve (AUC). Among these, S measures the model's highest accuracy for seen combinations when the calibration bias is set to +$\infty$, while U reflects the highest accuracy for unseen combinations when the bias is set to -$\infty$. HM represents the point where the model achieves the optimal balance between the prediction accuracies of seen and unseen categories. AUC, computed by dynamically adjusting the bias range from -$\infty$ to +$\infty$, represents the area under the curve of seen versus unseen accuracy. As a consequence, AUC serves as the core metric that can best reflects the overall performance of the model.

\subsubsection{Implementation Details}
We implement the proposed model based on PyTorch, using CLIP's backbone with the ViT-L/14 architecture, fine-tuned via Low-Rank Adaptation (LoRA) \cite{r28}. All experiments are conducted on an Nvidia H20 GPU. During training, we use the Adam optimizer, combined with learning rate decay and weight decay strategies. To simplify the model complexity, we use only one cross-attention layer for both local feature interaction and global feature fusion across the three datasets, with 12 attention heads and a dropout rate set to 0.1. The parameter \(\beta\), used to control the inference weights of each branch, is set to 0.1, 1.0 and 0.1 for MIT-States, UT-Zappos and C-GQA in the close-world setting, and set to 0.3, 1.0 and 0.3 in the open-world setting. Additionally, in the open-world setting, we introduce a feasibility score as a threshold to eliminate unreasonable combinations, effectively reducing the search space. The specific threshold is determined based on the model's performance on the validation set.

\subsection{Main Results}
We compare MSCI with other CZSL models using the same backbone (ViT-L/14). This comparison includes both CLIP-based and non-CLIP-based models. The results in the close-world setting are shown in Table \ref{tab:2}, while the results in the open world setting are presented in Table \ref{tab:3}, respectively. 

In the close-world setting, MSCI achieves optimal AUC and H metrics in all three datasets, with AUC improvements of 1.8\%, 9.8\%, and 14.5\% in MIT-States, UT-Zappos and C-GQA, respectively, compared to the second-best models. Such improvement percentages are closely related to the level of fine-grained information contained in the datasets: compared to MIT-States, UT-Zappos and C-GQA include richer fine-grained details, resulting in more significant performance improvements.

In the open-world setting, MSCI continues to exhibit exceptional performance, with AUC improvements of 13.0\% and 40.7\% on UT-Zappos and C-GQA, respectively. The greater performance improvements gained in the open-world setting can be attributed to the expanded search space, where discriminative local information becomes increasingly critical during inference. MSCI effectively exploits this information, maintaining robust generalization capabilities and superior performance.

\begin{table*}[ht]
\centering
\resizebox{\textwidth}{!}{%
\begin{tabular}{lllcccccccccccccc}
\toprule
\textbf{Category} & \textbf{Model} & \textbf{Venue} & \multicolumn{4}{c}{\textbf{MIT-States}} & \multicolumn{4}{c}{\textbf{UT-Zappos}} & \multicolumn{4}{c}{\textbf{C-GQA}} \\
\cmidrule(r){4-7} \cmidrule(r){8-11} \cmidrule(r){12-15}
 &  &  & S & U & H & AUC & S & U & H & AUC & S & U & H & AUC \\
\midrule
&SCEN \cite{r19} & CVPR & 29.9 & 25.2 & 18.4 & 5.3 & 63.5 & 63.1 & 47.8 & 32.0& 28.9 & 25.4 & 17.5 & 5.5 \\
 &OADis \cite{r20} & CVPR & 31.1 & 25.6 & 18.9 & 5.9 & 59.5 & 65.5 & 44.4 & 30.0& 33.4 & 14.3 & 14.7 & 3.8 \\
non CLIP-based models &CANet \cite{r22} & CVPR & 29.0 & 26.2 & 17.9 & 5.4 & 61.0 & 66.3 & 47.3 & 33.1 & 30.0 & 13.2 & 14.5 & 3.4 \\
 &CAPE \cite{r23} & ICCV & 32.1 & 28.0 & 20.4 & 6.7 & 62.3 & 68.5 & 49.5 & 35.2 & 33.0 & 16.4 & 16.3 & 4.6 \\
 & ADE \cite{r21} & CVPR & - & - & - & - & 63.0 & 64.3 & 51.1 & 35.1 & 35.0 & 17.7 & 18.0 & 5.2 \\
\midrule
 & CSP \cite{r7}  & ICLR & 46.6 & 49.9 & 36.3 & 19.4 & 64.2 & 66.2 & 46.6 & 33.0 & 28.8 & 26.8 & 20.5 & 6.2 \\
 & DFSP \cite{r25}  & CVPR & 46.9 & 52.0 & 37.3 & 20.6 & 66.7 & 71.7 & 47.2 & 36.9 & 38.2 & 32.9 & 27.1 & 10.5 \\
 & HPL \cite{r31}  & IJCAI & 47.5 & 50.6 & 37.3 & 20.2 & 63.0 & 68.8 & 48.2 & 35.0 & 30.8 & 28.4 & 22.4 & 7.2 \\
 & GIPCOL \cite{r10}  & WACV & 48.5 & 49.6 & 36.6 & 19.9 & 65.0 & 68.5 & 48.8 & 36.2 & 31.9 & 28.4 & 22.5 & 7.1 \\
CLIP-based models & Troika \cite{r8} & CVPR & 49.0 & \underline{53.0} & \underline{39.3} & 22.1 & 66.8 & 73.8 & \underline{54.6} & \underline{41.7} &\underline{41.0} &\underline{35.7} & \underline{29.4} &\underline{12.4} \\
 & CDS-CZSL \cite{r24}  & CVPR & \textbf{50.3} & 52.9 & 39.2 &\underline{22.4} & 63.9 &\underline{74.8} & 52.7 & 39.5 & 38.3 & 34.2 & 28.1 & 11.1 \\
 &PLID \cite{r29} &ECCV&49.7&52.4&39.0&22.1&\underline{67.3}&68.8&52.4&38.7&38.8&33.0&27.9&11.0\\
 

\midrule
 & MSCI & IJCAI & \underline {50.2} & \textbf{53.4} & \textbf{39.9}&\textbf{22.8} & \textbf{67.4} &\textbf{ 75.5} & \textbf{59.2} & \textbf{45.8 }& \textbf{42.4 }& \textbf{38.2} & \textbf{31.7} & \textbf{14.2} \\
\bottomrule
\end{tabular}
}

\caption{Comparison with other models in the close-world setting. The best results are in \textbf{bold}, and the second-best results are \underline {underlined}. }
\label{tab:2}
\end{table*}

\begin{table*}[ht]
\centering
\resizebox{\textwidth}{!}{%
\begin{tabular}{lllcccccccccccccc}
\toprule
\textbf{Category} & \textbf{Model} & \textbf{Venue} & \multicolumn{4}{c}{\textbf{MIT-States}} & \multicolumn{4}{c}{\textbf{UT-Zappos}} & \multicolumn{4}{c}{\textbf{C-GQA}} \\
\cmidrule(r){4-7} \cmidrule(r){8-11} \cmidrule(r){12-15}
 &  &  & S & U & H & AUC & S & U & H & AUC & S & U & H & AUC \\
\midrule
 
&KG-SP \cite{r27}  & CVPR & 28.4 & 7.5 & 7.4 & 1.3 & 61.8 & 52.1 & 42.3 & 26.5 & 31.5 & 2.9 & 4.7 & 0.8 \\
non CLIP-based models &DRANet \cite{r26} & ICCV & 29.8 & 7.8 & 7.9 & 1.5 & 65.1 & 54.3 & 44.0 & 28.8 & 31.3 & 3.9 & 6.0 & 1.1 \\
 
 & ADE \cite{r21} & CVPR & - & - & - & - & 62.4 & 50.7 & 44.8 & 27.1 & 35.1 & 4.8 & 7.6 & 1.4 \\
\midrule
 & CSP \cite{r7}  & ICLR & 46.3 & 15.7 & 17.4 & 5.7 & 64.1 & 44.1 & 38.9 & 22.7 & 28.7 & 5.2 & 6.9 & 1.2 \\
 & DFSP \cite{r25} & CVPR & 47.5 & 18.5 & 19.3 & 6.8 & 66.8 & 60.0 & 44.0 & 30.3 & 38.3 & 7.2 & 10.4 & 2.4 \\
  & HPL \cite{r31}  & IJCAI & 46.4 & 18.9 & 19.8 & 6.9 & 63.4 & 48.1 & 40.2 & 24.6 & 30.1 & 5.8 & 7.5 & 1.4 \\
 & GIPCOL \cite{r10}  & WACV & 48.5 & 16.0 & 17.9 & 6.3 & 65.0 & 45.0 & 40.1 & 23.5 & 31.6 & 5.5 & 7.3 & 1.3 \\
CLIP-based models & Troika \cite{r8} & CVPR & 48.8 & 18.7 & 20.1 & 7.2 & 66.4 & 61.2 & 47.8 &\underline{33.0} & \underline{40.8} & 7.9 & 10.9 &\underline{2.7} \\
 & CDS-CZSL \cite{r24} & CVPR & \textbf{49.4} & \textbf{21.8} & \textbf{22.1} & \textbf{8.5} & 64.7 &\underline {61.3} &\underline{48.2} & 32.3 & 37.6 &\underline{8.2} &\underline{11.6} &\underline{2.7} \\
 &PLID \cite{r29}& ECCV&49.1&18.7&20.4&7.3&\textbf{67.6}&55.5&46.6&30.8&39.1&7.5&10.6&2.5\\

\midrule
 & MSCI &IJCAI &\underline {49.2} &\underline{20.6} &\underline{21.2} &\underline{7.9} & \underline {67.4} &\textbf{ 63.0} & \textbf{53.2} & \textbf{37.3 }& \textbf{42.0} &\textbf{10.6} & \textbf{13.7} & \textbf{3.8} \\
\bottomrule
\end{tabular}
}
\caption{Comparison with other models in the open-world setting. The best results are in \textbf{bold}, and the second-best results are \underline {underlined}.}
\label{tab:3}
\end{table*}

\subsection{Ablation Study}
To further validate the effectiveness of each module in MSCI, we conduct ablation experiments on the UT-zappos dataset. The results are shown in Table \ref{tab:4} .
\subsubsection{Ablation for Aggregator}
To validate the effectiveness of the multi-layer feature aggregation module, we replace it with the following two alternative approaches: First, by using the features from the first and last layers of the visual encoder to perform cross-modal interactions in the first and second stages, respectively($w/o \, \text{Agg}_a$). Second, by using the average of the features from the first N layers and the average of the features from the last N layers to perform stage-wise cross-modal interactions($w/o \, \text{Agg}_b$). The experimental results show that, compared to the two methods mentioned above, the proposed adaptive fusion module achieves better aggregation while maintaining information richness.

\subsubsection{Ablation for Multi-stage Cross-modal Interaction}
To validate the effectiveness of the multi-stage cross-modal interaction module, we remove the interaction modules in the first ($w/o \, \text{Ms}_a$) and second stages($w/o \, \text{Ms}_b$) in two separate ablation scenarios. The experimental results show that, compared to single-stage cross-modal interaction, stage-wise cross-modal interaction is able to incorporate global visual information into the prompt embedding and further integrate rich local visual information, thereby achieving better results.

\begin{table}[H]
\centering
\resizebox{0.35\textwidth}{!}{
\renewcommand{\arraystretch}{1.2} 
\begin{tabular}{lcccc}
\hline
\textbf{Ablation Experiment} & \textbf{S} & \textbf{U} & \textbf{H} & \textbf{AUC} \\ \hline
$w/o \, \text{Agg}_a$ & \textbf{68.1} & 74.0 & 54.3 & 42.6 \\
$w/o \, \text{Agg}_b$ & 65.9 & 75.2 & 56.2 & 43.2 \\
$w/o \, \text{Ms}_a$  & 63.8 & 70.4 & 51.5 & 37.4 \\
$w/o \, \text{Ms}_b$  & 66.8 & 75.2 & 56.1 & 43.1 \\
$w/o \, \text{Df}$    & 67.2 & \textbf{76.4} & \underline{57.3} & \underline{45.3} \\ \hline
 & \underline{67.4} & \underline{75.5} & \textbf{59.2} & \textbf{45.8} \rule{0pt}{2.5ex} \\ \hline
\end{tabular}
}
\caption{Ablation results for UT-Zappos in the close-world setting.}
\label{tab:4}
\end{table}

\subsubsection{Ablation for Dynamic Fusion}
To validate the effectiveness of the proposed fusion method, we replace it with the direct use of the output from the multi-stage cross-modal interaction module (i.e., removing the $\lambda_1 t_1$ term in Equation 9) ($w/o \, \text{Df}$). Based on the experimental results, compared to the single fusion approach, the fusion method we propose can dynamically adjust the attention to global and local visual information according to different combinations, as well as the different prompt branches of the same combination, achieving superior performance.

\subsection{Qualitative Results}

\begin{figure}[ht] 
    \centering 
    \includegraphics[width=0.5\textwidth]{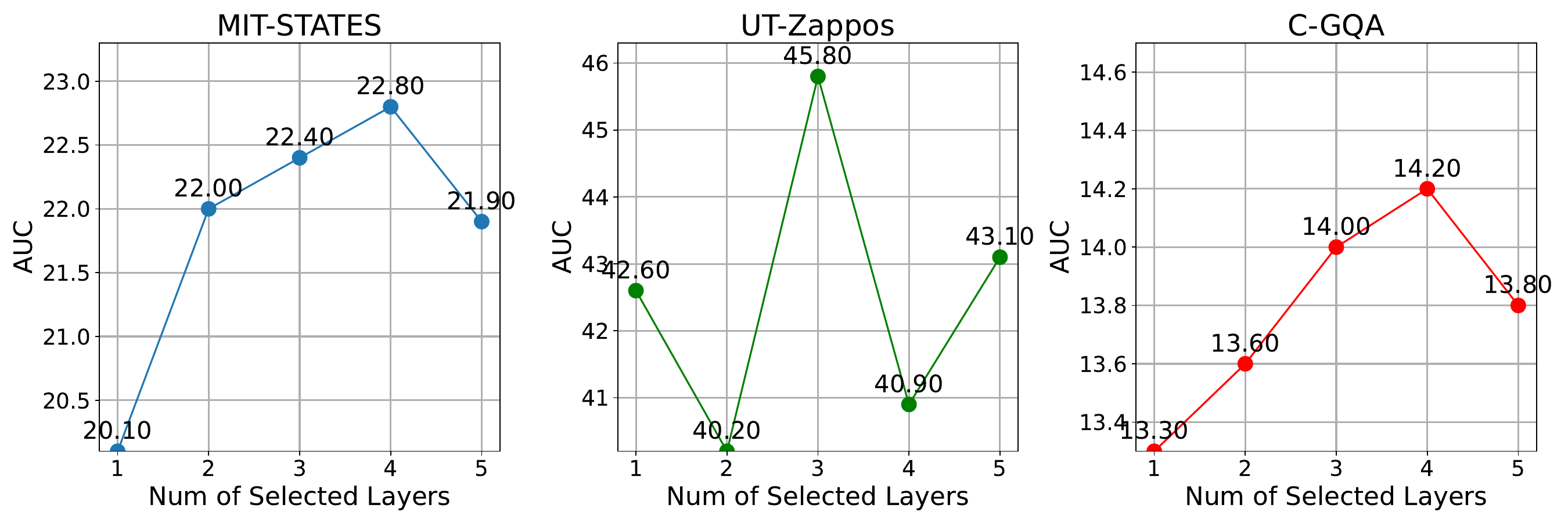} 
    \caption{The impact of the number of aggregated layers \(N\) on AUC.} 
    \label{fig:4} 
\end{figure}

Within the framework of MSCI, we adjust the number of selected layers based on the complexity of each dataset. For the relatively simple UT-zappos dataset, selecting features from the first three and the last three layers for aggregation has proven to optimize model performance. In contrast, for the more structurally complex MiT-States and C-GQA datasets, processing features from the first four and the last four layers is more effective to ensure optimal results. The variation in the AUC metric with the number of selected layers \(N\) for each dataset is shown in Figure \ref{fig:4}. Notably, extensive evaluations suggest that setting \(M\) and \(N\) to identical values yields better performance; thus, this configuration is adopted by default.

\begin{figure}[ht] 
    \centering 
    \includegraphics[width=0.40\textwidth]{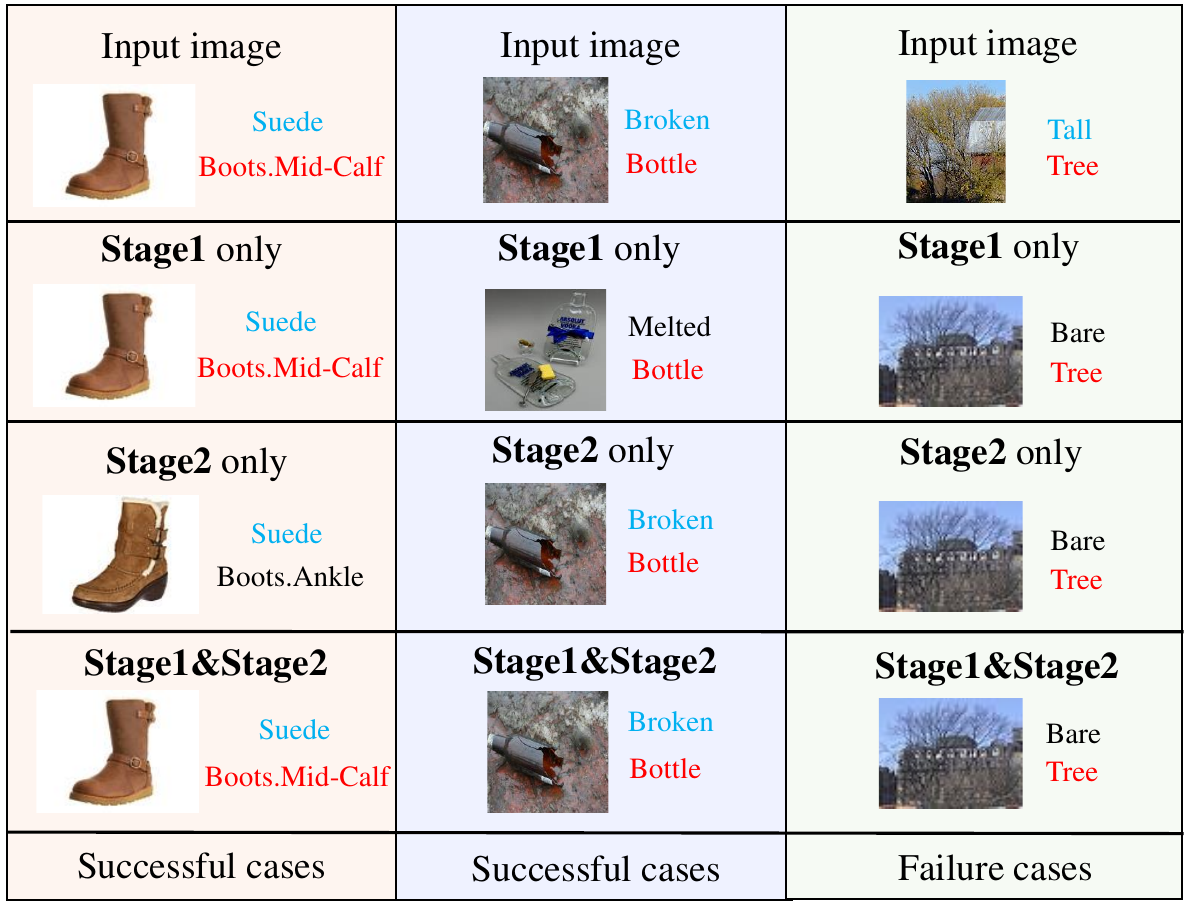} 
    \caption{Qualitative analysis of the proposed model.} 
    \label{fig:5} 
\end{figure}

In addition, we select one typical case from each of the three datasets for qualitative analysis, as shown in Figure \ref{fig:5}. Through the two successful cases, it can be intuitively observed that, compared to the single-stage interaction model, MSCI's multi-stage interaction can more effectively and comprehensively integrate cross-modal information, demonstrating significant advantages. However, in the failure case, we observe that certain distractors were highly similar to the actual items, causing MSCI to misjudge both local and global features, which further led to incorrect final prediction results.

\section{Conclusion}

In this study, we propose a novel model called MSCI for compositional zero-shot learning. MSCI employs an adaptive mechanism to progressively integrate local and global information from lower and higher visual feature layers and incorporate them into the prompt embeddings in a stage-by-stage manner, which effectively overcomes CLIP's inherent limitations for capturing local visual details. Furthermore, MSCI can autonomously optimize the allocation of attention weights on local details and global visual information based on different combinations, as well as the different elements within the same combination. Experiments show that MSCI achieves significant improvements in terms of various evaluation metrics on three widely used datasets.

\newpage
\section*{Acknowledgments}
This work is supported by the Natural Science Foundation of China under Grant No. 62302213, the Key Laboratory of Social Computing and Cognitive Intelligence (Dalian University of Technology), Ministry of Education, China, and by the Postgraduate Research and Practice Innovation Program of Jiangsu Province under Grant No. SJCX25\_0162.
\bibliographystyle{named}
\bibliography{ijcai25}

\begin{thebibliography}{}

\bibitem[\protect\citeauthoryear{Bao \bgroup \em et al.\egroup }{2025}]{r29}
Wentao Bao, Lichang Chen, Heng Huang, and Yu~Kong.
\newblock Prompting language-informed distribution for compositional zero-shot learning.
\newblock In {\em European Conference on Computer Vision}, pages 107--123. Springer, 2025.

\bibitem[\protect\citeauthoryear{Du \bgroup \em et al.\egroup }{2021}]{r34}
Jinlong Du, Senzhang Wang, Hao Miao, and Jiaqiang Zhang.
\newblock Multi-channel pooling graph neural networks.
\newblock In {\em IJCAI}, pages 1442--1448, 2021.

\bibitem[\protect\citeauthoryear{Hao \bgroup \em et al.\egroup }{2023}]{r21}
Shaozhe Hao, Kai Han, and Kwan-Yee~K Wong.
\newblock Learning attention as disentangler for compositional zero-shot learning.
\newblock In {\em Proceedings of the IEEE/CVF Conference on Computer Vision and Pattern Recognition}, pages 15315--15324, 2023.

\bibitem[\protect\citeauthoryear{Hu \bgroup \em et al.\egroup }{2021}]{r28}
Edward~J Hu, Yelong Shen, Phillip Wallis, Zeyuan Allen-Zhu, Yuanzhi Li, Shean Wang, Lu~Wang, and Weizhu Chen.
\newblock Lora: Low-rank adaptation of large language models.
\newblock {\em arXiv preprint arXiv:2106.09685}, 2021.

\bibitem[\protect\citeauthoryear{Huang \bgroup \em et al.\egroup }{2024}]{r8}
Siteng Huang, Biao Gong, Yutong Feng, Min Zhang, Yiliang Lv, and Donglin Wang.
\newblock Troika: Multi-path cross-modal traction for compositional zero-shot learning.
\newblock In {\em Proceedings of the IEEE/CVF Conference on Computer Vision and Pattern Recognition}, pages 24005--24014, 2024.

\bibitem[\protect\citeauthoryear{Hudson and Manning}{2019}]{r33}
Drew~A Hudson and Christopher~D Manning.
\newblock Gqa: A new dataset for real-world visual reasoning and compositional question answering.
\newblock In {\em Proceedings of the IEEE/CVF conference on computer vision and pattern recognition}, pages 6700--6709, 2019.

\bibitem[\protect\citeauthoryear{Isola \bgroup \em et al.\egroup }{2015}]{r16}
Phillip Isola, Joseph~J Lim, and Edward~H Adelson.
\newblock Discovering states and transformations in image collections.
\newblock In {\em Proceedings of the IEEE Conference on Computer Vision and Pattern Recognition}, pages 1383--1391, 2015.

\bibitem[\protect\citeauthoryear{Jing \bgroup \em et al.\egroup }{2024}]{r9}
Chenchen Jing, Yukun Li, Hao Chen, and Chunhua Shen.
\newblock Retrieval-augmented primitive representations for compositional zero-shot learning.
\newblock In {\em Proceedings of the AAAI Conference on Artificial Intelligence}, volume~38, pages 2652--2660, 2024.

\bibitem[\protect\citeauthoryear{Karthik \bgroup \em et al.\egroup }{2022}]{r27}
Shyamgopal Karthik, Massimiliano Mancini, and Zeynep Akata.
\newblock Kg-sp: Knowledge guided simple primitives for open world compositional zero-shot learning.
\newblock In {\em Proceedings of the IEEE/CVF Conference on Computer Vision and Pattern Recognition}, pages 9336--9345, 2022.

\bibitem[\protect\citeauthoryear{Khan \bgroup \em et al.\egroup }{2023}]{r23}
Muhammad Gul Zain~Ali Khan, Muhammad~Ferjad Naeem, Luc Van~Gool, Alain Pagani, Didier Stricker, and Muhammad~Zeshan Afzal.
\newblock Learning attention propagation for compositional zero-shot learning.
\newblock In {\em Proceedings of the IEEE/CVF Winter Conference on Applications of Computer Vision}, pages 3828--3837, 2023.

\bibitem[\protect\citeauthoryear{Kruse \bgroup \em et al.\egroup }{2022}]{r30}
Rudolf Kruse, Sanaz Mostaghim, Christian Borgelt, Christian Braune, and Matthias Steinbrecher.
\newblock Multi-layer perceptrons.
\newblock In {\em Computational intelligence: a methodological introduction}, pages 53--124. Springer, 2022.

\bibitem[\protect\citeauthoryear{Li \bgroup \em et al.\egroup }{2022}]{r19}
Xiangyu Li, Xu~Yang, Kun Wei, Cheng Deng, and Muli Yang.
\newblock Siamese contrastive embedding network for compositional zero-shot learning.
\newblock In {\em Proceedings of the IEEE/CVF Conference on Computer Vision and Pattern Recognition}, pages 9326--9335, 2022.

\bibitem[\protect\citeauthoryear{Li \bgroup \em et al.\egroup }{2023}]{r26}
Yun Li, Zhe Liu, Saurav Jha, and Lina Yao.
\newblock Distilled reverse attention network for open-world compositional zero-shot learning.
\newblock In {\em Proceedings of the IEEE/CVF International Conference on Computer Vision}, pages 1782--1791, 2023.

\bibitem[\protect\citeauthoryear{Li \bgroup \em et al.\egroup }{2024a}]{r38}
Yicong Li, Xiangguo Sun, Hongxu Chen, Sixiao Zhang, Yu~Yang, and Guandong Xu.
\newblock Attention is not the only choice: Counterfactual reasoning for path-based explainable recommendation.
\newblock {\em IEEE Transactions on Knowledge and Data Engineering}, 2024.

\bibitem[\protect\citeauthoryear{Li \bgroup \em et al.\egroup }{2024b}]{r37}
Yicong Li, Yu~Yang, Jiannong Cao, Shuaiqi Liu, Haoran Tang, and Guandong Xu.
\newblock Toward structure fairness in dynamic graph embedding: A trend-aware dual debiasing approach.
\newblock In {\em Proceedings of the 30th ACM SIGKDD Conference on Knowledge Discovery and Data Mining}, pages 1701--1712, 2024.

\bibitem[\protect\citeauthoryear{Li \bgroup \em et al.\egroup }{2024c}]{r24}
Yun Li, Zhe Liu, Hang Chen, and Lina Yao.
\newblock Context-based and diversity-driven specificity in compositional zero-shot learning.
\newblock In {\em Proceedings of the IEEE/CVF Conference on Computer Vision and Pattern Recognition}, pages 17037--17046, 2024.

\bibitem[\protect\citeauthoryear{Liu \bgroup \em et al.\egroup }{2024}]{r13}
Ruitong Liu, Yanbin Wang, Zhenhao Guo, Haitao Xu, Zhan Qin, Wenrui Ma, and Fan Zhang.
\newblock Transurl: Improving malicious url detection with multi-layer transformer encoding and multi-scale pyramid features.
\newblock {\em Computer Networks}, 253:110707, 2024.

\bibitem[\protect\citeauthoryear{Lu \bgroup \em et al.\egroup }{2023}]{r25}
Xiaocheng Lu, Song Guo, Ziming Liu, and Jingcai Guo.
\newblock Decomposed soft prompt guided fusion enhancing for compositional zero-shot learning.
\newblock In {\em Proceedings of the IEEE/CVF Conference on Computer Vision and Pattern Recognition}, pages 23560--23569, 2023.

\bibitem[\protect\citeauthoryear{Miao \bgroup \em et al.\egroup }{2025}]{r35}
Hao Miao, Ronghui Xu, Yan Zhao, Senzhang Wang, Jianxin Wang, Philip~S Yu, and Christian~S Jensen.
\newblock A parameter-efficient federated framework for streaming time series anomaly detection via lightweight adaptation.
\newblock {\em TMC}, 2025.

\bibitem[\protect\citeauthoryear{Misra \bgroup \em et al.\egroup }{2017}]{r32}
Ishan Misra, Abhinav Gupta, and Martial Hebert.
\newblock From red wine to red tomato: Composition with context.
\newblock In {\em Proceedings of the IEEE Conference on Computer Vision and Pattern Recognition}, pages 1792--1801, 2017.

\bibitem[\protect\citeauthoryear{Naeem \bgroup \em et al.\egroup }{2021}]{r2}
Muhammad~Ferjad Naeem, Yongqin Xian, Federico Tombari, and Zeynep Akata.
\newblock Learning graph embeddings for compositional zero-shot learning.
\newblock In {\em Proceedings of the IEEE/CVF Conference on Computer Vision and Pattern Recognition}, pages 953--962, 2021.

\bibitem[\protect\citeauthoryear{Nagarajan and Grauman}{2018}]{r3}
Tushar Nagarajan and Kristen Grauman.
\newblock Attributes as operators: factorizing unseen attribute-object compositions.
\newblock In {\em Proceedings of the European Conference on Computer Vision (ECCV)}, pages 169--185, 2018.

\bibitem[\protect\citeauthoryear{Nan \bgroup \em et al.\egroup }{2019}]{r4}
Zhixiong Nan, Yang Liu, Nanning Zheng, and Song-Chun Zhu.
\newblock Recognizing unseen attribute-object pair with generative model.
\newblock In {\em Proceedings of the AAAI Conference on Artificial Intelligence}, volume~33, pages 8811--8818, 2019.

\bibitem[\protect\citeauthoryear{Nayak \bgroup \em et al.\egroup }{2022}]{r7}
Nihal~V Nayak, Peilin Yu, and Stephen~H Bach.
\newblock Learning to compose soft prompts for compositional zero-shot learning.
\newblock {\em arXiv preprint arXiv:2204.03574}, 2022.

\bibitem[\protect\citeauthoryear{Purushwalkam \bgroup \em et al.\egroup }{2019}]{r1}
Senthil Purushwalkam, Maximilian Nickel, Abhinav Gupta, and Marc'Aurelio Ranzato.
\newblock Task-driven modular networks for zero-shot compositional learning.
\newblock In {\em Proceedings of the IEEE/CVF International Conference on Computer Vision}, pages 3593--3602, 2019.

\bibitem[\protect\citeauthoryear{Radford \bgroup \em et al.\egroup }{2021}]{r5}
Alec Radford, Jong~Wook Kim, Chris Hallacy, Aditya Ramesh, Gabriel Goh, Sandhini Agarwal, Girish Sastry, Amanda Askell, Pamela Mishkin, Jack Clark, et~al.
\newblock Learning transferable visual models from natural language supervision.
\newblock In {\em International Conference on Machine Learning}, pages 8748--8763. PMLR, 2021.

\bibitem[\protect\citeauthoryear{Saini \bgroup \em et al.\egroup }{2022}]{r20}
Nirat Saini, Khoi Pham, and Abhinav Shrivastava.
\newblock Disentangling visual embeddings for attributes and objects.
\newblock In {\em Proceedings of the IEEE/CVF Conference on Computer Vision and Pattern Recognition}, pages 13658--13667, 2022.

\bibitem[\protect\citeauthoryear{Scarselli \bgroup \em et al.\egroup }{2008}]{r11}
Franco Scarselli, Marco Gori, Ah~Chung Tsoi, Markus Hagenbuchner, and Gabriele Monfardini.
\newblock The graph neural network model.
\newblock {\em IEEE transactions on neural networks}, 20(1):61--80, 2008.

\bibitem[\protect\citeauthoryear{Tang \bgroup \em et al.\egroup }{2023}]{r12}
Feilong Tang, Zhongxing Xu, Qiming Huang, Jinfeng Wang, Xianxu Hou, Jionglong Su, and Jingxin Liu.
\newblock Duat: Dual-aggregation transformer network for medical image segmentation.
\newblock In {\em Chinese Conference on Pattern Recognition and Computer Vision (PRCV)}, pages 343--356. Springer, 2023.

\bibitem[\protect\citeauthoryear{Wang \bgroup \em et al.\egroup }{2022}]{r15}
Ruotong Wang, Yanqing Shen, Weiliang Zuo, Sanping Zhou, and Nanning Zheng.
\newblock Transvpr: Transformer-based place recognition with multi-level attention aggregation.
\newblock In {\em Proceedings of the IEEE/CVF Conference on Computer Vision and Pattern Recognition}, pages 13648--13657, 2022.

\bibitem[\protect\citeauthoryear{Wang \bgroup \em et al.\egroup }{2023a}]{r31}
Henan Wang, Muli Yang, Kun Wei, and Cheng Deng.
\newblock Hierarchical prompt learning for compositional zero-shot recognition.
\newblock In {\em IJCAI}, volume~1, page~3, 2023.

\bibitem[\protect\citeauthoryear{Wang \bgroup \em et al.\egroup }{2023b}]{r22}
Qingsheng Wang, Lingqiao Liu, Chenchen Jing, Hao Chen, Guoqiang Liang, Peng Wang, and Chunhua Shen.
\newblock Learning conditional attributes for compositional zero-shot learning.
\newblock In {\em Proceedings of the IEEE/CVF Conference on Computer Vision and Pattern Recognition}, pages 11197--11206, 2023.

\bibitem[\protect\citeauthoryear{Xu \bgroup \em et al.\egroup }{2024a}]{r10}
Guangyue Xu, Joyce Chai, and Parisa Kordjamshidi.
\newblock Gipcol: Graph-injected soft prompting for compositional zero-shot learning.
\newblock In {\em Proceedings of the IEEE/CVF Winter Conference on Applications of Computer Vision}, pages 5774--5783, 2024.

\bibitem[\protect\citeauthoryear{Xu \bgroup \em et al.\egroup }{2024b}]{r36}
Ronghui Xu, Hao Miao, Senzhang Wang, Philip~S Yu, and Jianxin Wang.
\newblock Pefad: a parameter-efficient federated framework for time series anomaly detection.
\newblock In {\em SIGKDD}, pages 3621--3632, 2024.

\bibitem[\protect\citeauthoryear{Yang \bgroup \em et al.\egroup }{2023}]{r14}
Qu~Yang, Mang Ye, Zhaohui Cai, Kehua Su, and Bo~Du.
\newblock Composed image retrieval via cross relation network with hierarchical aggregation transformer.
\newblock {\em IEEE Transactions on Image Processing}, 32, 2023.

\bibitem[\protect\citeauthoryear{Yu and Grauman}{2014}]{r17}
Aron Yu and Kristen Grauman.
\newblock Fine-grained visual comparisons with local learning.
\newblock In {\em Proceedings of the IEEE Conference on Computer Vision and Pattern Recognition}, pages 192--199, 2014.

\bibitem[\protect\citeauthoryear{Zhou \bgroup \em et al.\egroup }{2022}]{r6}
Kaiyang Zhou, Jingkang Yang, Chen~Change Loy, and Ziwei Liu.
\newblock Learning to prompt for vision-language models.
\newblock {\em International Journal of Computer Vision}, page 2337–2348, Sep 2022.

\end{thebibliography}

\end{document}